%% file: sample-sigconf.tex
\documentclass[sigconf]{acmart}
\renewcommand\footnotetextcopyrightpermission[1]{} 
\usepackage{booktabs} 

\usepackage{mathtools}
\usepackage{graphicx}
\usepackage{amsmath,amssymb} 
\usepackage{color}
\usepackage{subcaption}
\usepackage{url}
\usepackage{balance} 
\usepackage{float}

\begin{document}

\title{Recovering Facial Reflectance and Geometry \\ from Multi-view Images}


\author{Guoxian Song}
\authornote{The Corresponding author.}
\affiliation{%
	\institution{Nanyang Technological University}
}
\email{guoxian001@e.ntu.edu.sg}

\author{Jianmin Zheng}
\affiliation{%
	\institution{Nanyang Technological University}
}
\email{ASJMZheng@ntu.edu.sg}

\author{Jianfei Cai}
\affiliation{%
	\institution{Nanyang Technological University}
}
\email{asjfcai@ntu.edu.sg}

\author{Tat-Jen Cham}
\affiliation{%
	\institution{Nanyang Technological University}
}
\email{astjcham@ntu.edu.sg}

\begin{abstract}
While the problem of estimating shapes and diffuse reflectances of human faces from images  has been extensively studied, there is relatively less work done on recovering the specular albedo. This paper presents a {\em lightweight} solution for inferring photorealistic facial reflectance and geometry. Our system processes video streams from two views of a subject, and outputs two reflectance maps for diffuse and specular albedos, as well as a vector map of surface normals. A model-based optimization approach is used, consisting of the three stages of multi-view face model fitting, facial reflectance inference and facial geometry refinement. Our approach is based on a novel formulation built upon the 3D morphable model (3DMM) for representing 3D textured faces in conjunction with the Blinn-Phong reflection model. It has the advantage of requiring only a simple setup with two video streams, and is able to exploit the interaction between the diffuse and specular reflections across multiple views as well as time frames. As a result, the method is able to reliably recover high-fidelity facial reflectance and geometry, which facilitates various applications such as generating photorealistic facial images under new viewpoints or illumination conditions.
\end{abstract}



\maketitle

\input{introduction}
\input{approach}
\input{experiment}

\section{Conclusions}
We have described a model based method to recover photorealistic facial reflectance and geometry from two video streams of a subject in two views. By jointly estimating initial facial geometry and texture map, jointly inferring specular and diffuse reflectance, and further refining geometry while utilizing two view information, our approach demonstrates appreciable improvements over the prior art. It allows for better reconstruction of the shape of faces with specular reflections and more compelling rendering of faces with specular effects under new viewpoints. 

Our approach has some limitations. For example, we use an iterative optimization approach, which is not as fast as single-pass network-based approaches. However, the data generated by our method can be used as pseudo ground truth for improving the training of such network-based methods. In addition, our method assumes a consistent lighting environment.  For dynamic lighting, we will need to re-estimate the lighting condition for each frame. Furthermore, our approach may not accurately recover unmodeled details such as kerchief and hair, which do not correspond to semantic features of human faces and may interfere with the fitting process used to recover corresponding texture maps.

\section{Acknowledgments}
We would like to thank Yudong Guo and Juyong Zhang from University of Science and Technology of China for providing comparison result. And we also thank Chuanxia Zhen, Yujun Cai, Zhijie Zhang, Ayan Kumar Bhunia, etc for data collection. This research is supported by Singtel Cognitive and Artificial Intelligence Lab for Enterprises at NTU.

\bibliographystyle{ACM-Reference-Format}
\balance
\bibliography{sample-bibliography}

\end{document}

%% file: introduction.tex
\section{Introduction}
This paper considers the problem of recovering facial reflectance and geometry from images with emphasis on inferring specular albedo, which is important as real faces often exhibit bright highlights due to specular reflection. Accurately recovering specular components facilitates more realistic rendering of digital faces in new environments, which is very useful in many applications such as augmented and virtual reality, games, entertainment~\cite{Hu:2017:ADS:3130800.31310887,Song:2018:RFP:3240508.3240570,olszewski2016high} and telepresence~\cite{thies2016face,li2015facial,Garrido:2016,Cao:2015}. It also helps in the recovery of high-fidelity 3D geometry for faces, which is a fundamental problem in computer vision and graphics.

\begin{figure}[h]
	\centering
	\includegraphics[width=1.0\linewidth]{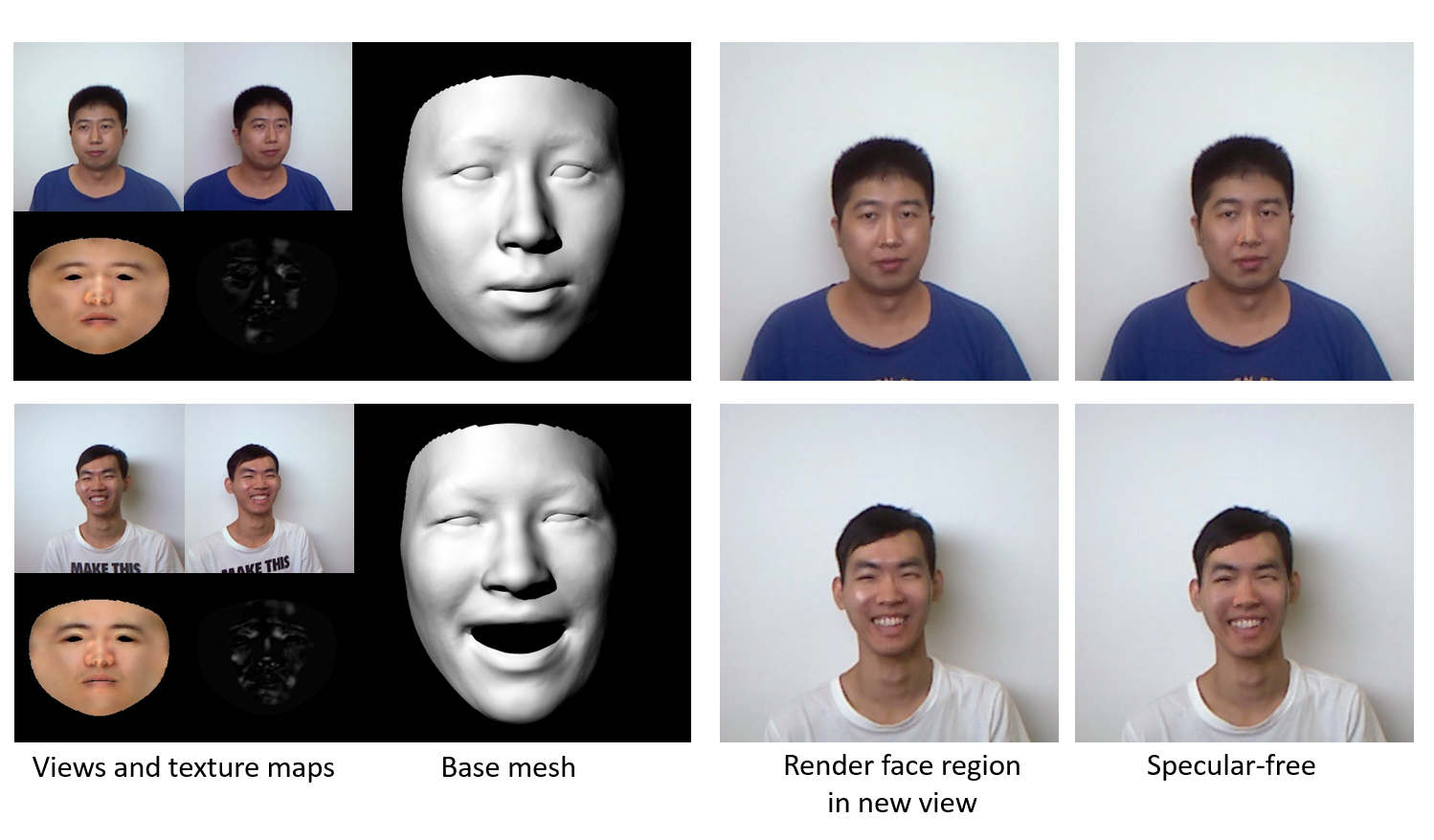}
	\caption{Given image streams from two views, our method can infer the 3D geometry as well as the diffuse and specular albedos of faces, together with the environment lighting. This allows rendering to new views with or without specular reflections.}
	\label{fig:Intro}
\end{figure}

Modeling high quality 3D geometry and reflectance from images is a very challenging task due to the complex yet ambiguous interaction across facial geometry, reflectance properties, illumination and camera parameters. Great success has been achieved via sophisticated systems that require specialized hardware, such as appearance measurement devices, in studio environments where the lighting can be fully controlled~\cite{Debevec:2000:ARF:344779.344855,Ma:2007:RAS:2383847.2383873,Ghosh:2008:PMA:1409060.1409092}. However, these professional configurations are cumbersome and costly. In contrast, various shading-from-X algorithms~\cite{thies2016face,Guo20173DFaceNet} requiring less stringent conditions may alternatively be used. Due to the highly nonlinear effects of the specular component, most of these methods assume approximately Lambertian surfaces and thus ignore specular reflection, which is detrimental to the accuracy of the modeling.

There are some approaches that handle non-Lambertian objects, but many of these have very limiting assumptions, or focus only on removing specularities. For example, Kumar et al.~\cite{kumar2011non} proposed a method for reconstructing human faces that takes both diffuse and specular reflections into account. Magda and Zickler proposed two approaches to reconstruct surfaces with arbitrary BRDFs~\cite{magda2001beyond}. However, both works require constraints on the locations of point light sources, or a large collection of images generated from a fixed view point with a single point light source.
There are works to infer facial albedo or geometry based on spherical harmonics lighting. For examples, Saito et al.\cite{DBLP:journals/corr/SaitoWHNL16} proposed to infer photorealistic texture details from a high resolution face texture database based on a deep neural network. However, it does not reconstruct fine-scale geometric detail for photorealistic rendering. Guo et al.~\cite{Guo20173DFaceNet} proposed real-time CNN-based 3D face capture from an RGB camera in a coarse-to-fine manner, which uses a deep neural network to regress a coarse shape and refine the details based on the input image. Recently, Yamaguchi et al.~\cite{Yamaguchi:2018:HFR:3197517.3201364} proposed a learning-based technique to infer high-quality facial reflectance and geometry from a single unconstrained image. It acquired training data by employing 7 high-resolution DSLR cameras and used polarized gradient spherical illumination to obtain diffuse and specular albedos, as well as medium- and high-frequency geometric displacement maps, but this setup demands high effort and cost.

Our work is inspired by the success of \cite{Guo20173DFaceNet} and \cite{Yamaguchi:2018:HFR:3197517.3201364}. However, different from \cite{Guo20173DFaceNet} that relies on PCA-based diffuse albedo estimation and view-based geometric refinement, our method  refines diffuse albedo and estimate specular albedo from input images via the Blinn-Phong model, and we also introduce a geometric normal refinement method to improve  the 3D normal displacement map. Instead of expensively collecting high-quality training data in \cite{Yamaguchi:2018:HFR:3197517.3201364} using a sophisticated multi-view system, our method will be more widely applicable to cost-sensitive scenarios, and does not require prior data collection. 

Diffuse and specular reflections behave differently under viewpoint changes.
As specular reflection depends on the angle of the surface normal to the camera but diffuse reflection does not, we propose to take frames from video streams to separate the effects caused by the diffuse albedo and the surface normal. Moreover, we build our formulation upon the 3D morphable model (3DMM), considering that this PCA (principal component analysis) based facial parametric model provides a compact representation for 3D faces and encodes some intrinsic structures of the faces. By non-trivially integrating all these new ideas, we provide a reliable and lightweight solution to recovering facial reflectance (including diffuse albedo and specular albedo) and geometry from two-view images. In addition, we collected a multi-view facial expression dataset of 15 subjects from different countries with different skin conditions for evaluation. The experiments demonstrate that 
our approach can substantially improve face recovery fidelity and produce photorealistic rendering at new viewpoints (See Figure \ref{fig:Intro}).

The main contributions of the paper are twofold:
\begin{itemize}
	\item We propose a reliable and lightweight solution for faithfully recovering facial appearances with specularity, which combines multi-view face model fitting, facial reflectance inference and geometry refinement. The solution is built on the 3D morphable model (3DMM) for representing 3D textured faces, and uses a simple setup of two-view video streams as input for separating the behavior of diffuse and specular reflections across multiple views and time frames. 

	\item We propose two optimization algorithms to recover the specular albedo based on the Blinn-Phong reflection model and to refine the normal map for surfaces generated by PCA based methods. Our idea of using the Blinn-Phong model for specular faces can be incorporated into many existing Lambertian based 3D facial modeling works and applications. In addition, we collected a test dataset with video streams of 15 subjects with different ethnicities, which will be released.
	
\end{itemize}

\section{Related Work}
This section briefly reviews some relevant work.

\textbf{Capturing facial appearance.}
Capturing facial appearance including detailed geometry and reflectance is of great interest in both academia and industry, and has been studied extensively. In particular, various sophisticates systems have been designed. For example, Debevec et al. developed the Light Stage technique, which can capture fine facial appearances~\cite{Debevec:2000:ARF:344779.344855}. The technique was improved by using specialized devices such as camera matrix and sphere LEDs~\cite{Ma:2007:RAS:2383847.2383873,Ghosh:2008:PMA:1409060.1409092,Ghosh:2011:MFC:2070781.2024163}, which
provides production-level measurement of reflectance. Recently, there is an effort to design deployable and portable systems for photorealistic facial capturing. Tan et al.~\cite{Tan:2017} introduced a cost-efficient framework based on two RGB-D inputs with depth sensors such as Kinect to align point clouds from two cameras for high quality 3D facial reconstruction. Our method simply needs two color webcams.

\textbf{Parametric facial models.}
There has also been a lot of research done on representing human faces using parametric models. 
Earliest works include Eigenfaces \cite{eigenface} and the active appearance model (AAM) \cite{Edwards:1998:IFI:520809.796067}, which represent facial appearance using linear models and facilitate many applications. Smith et.al proposed several works focusing on recovering facial shape using surface normal from a parametric model\cite{1717453,Smith:2008:FSR:1325290.1325297}. A well-known and widely-used model is 3D morphable model (3DMM) proposed by Blanz and Vetter \cite{Blanz:1999:MMS:311535.311556}, which models textured 3D faces by analysis-by-synthesis and  disentangles a 3D face into three sets of parameters: facial identity, expression and albedo. The basic idea of 3DMM leverages Principal Component Analysis (PCA) and transforms the shape and texture of example faces into a vector representation. 3DMM can be computed efficiently, but it is limited by the linear space spanned by the training samples and fails to capture fine geometric and appearance details.

\textbf{Reflection models.}
Synthesizing an image of face requires reflection modeling, which is used to determine the colour in a particular view. Reflectance models usually contain different reflection components such as diffuse colour, specular colour and ambient colour. To separate each component from an input image is a well-studied problem.  Early approaches \cite{Bajcsy1996DetectionOD} recover diffuse and specular colours by analysis of colour histogram distributions, under the piecewise-constant surface colour assumption. A recent approach \cite{Tan:2005:SRC:1038062.1038236} is to derive a pseudo diffuse image that exhibits the same geometric profile as the diffuse component of the input image; and then iteratively remove highlights and propagate the maximum diffuse component to neighbor pixels. Some approaches \cite{Kim2013SpecularRS} \cite{Yang:2010:RSH:1888089.1888097} employ a dark channel prior in generating the pseudo diffuse images. 

\textbf{Learning-based reflectance recovery.}
Recently learning-based techniques have been developed to infer facial geometry and colour properties from images. For example, Guo et al.~\cite{Guo20173DFaceNet} synthesized large datasets and trained a coarse-to-fine CNN framework for real-time 3D face reconstruction based on 3DMM with RGB inputs. Yu et al.~\cite{DBLP:journals/corr/abs-1709-00536} presented a neural network for dense facial correspondences in highly unconstrained RGB images. These works achieve impressive results on facial geometric fitting or reconstruction, but their reflectance shading assumption is Lambertian. Our work extends those methods to consider specular effect, which leads to more photorealistic rendering. Yamaguchi et al.~\cite{Yamaguchi:2018:HFR:3197517.3201364} used a deep learning approach to infer high-fidelity facial reflectance and geometry from a single image based on a large-scale dataset with ground truth, which is costly to obtain. 

%% file: approach.tex
\section{Overview}

\begin{figure*}[ht!]
	\includegraphics[width=1\linewidth]{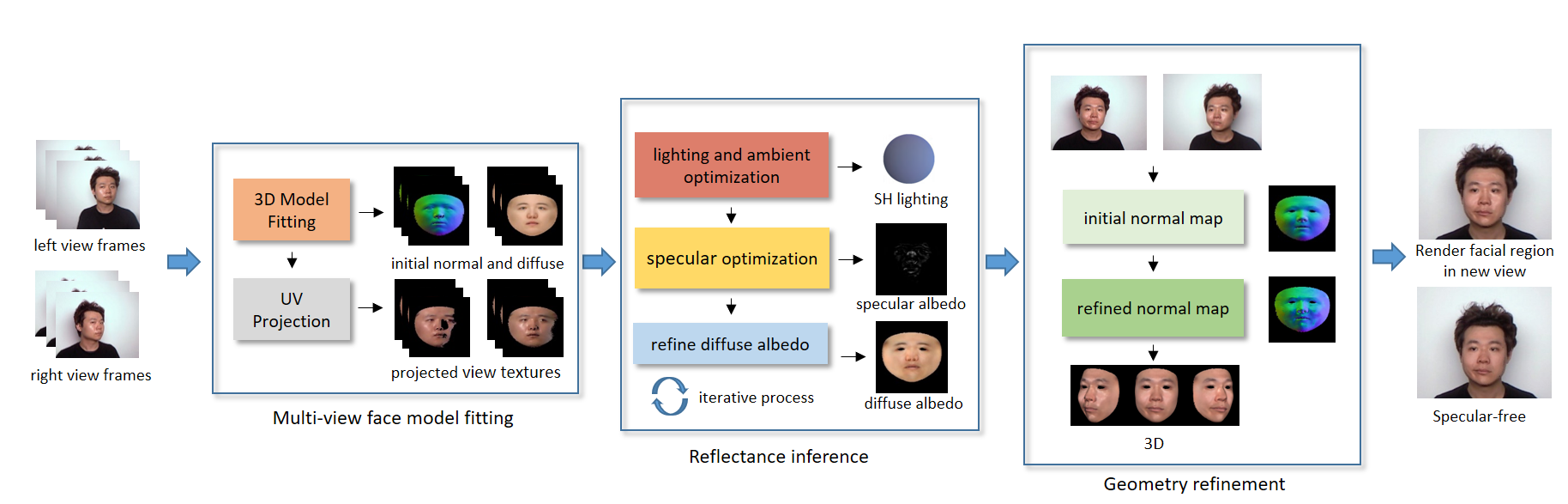}
	\caption{The pipeline of our approach.}
	\label{fig:overview}
\end{figure*}

The pipeline of our approach is illustrated in Figure~\ref{fig:overview}. The whole process consists of three stages shown from left to right, respectively. The first stage is multi-view 3D face model fitting, which fits a 3D face model to  two streaming views. We adapt an inverse rendering based optimization method \cite{Guo20173DFaceNet,thies2016face} to generate an initial diffuse albedo and normal map for each time instance. The second stage is facial reflectance inference, which estimates lighting and extracts the specular albedo map and refined diffuse albedo map iteratively, by minimizing image differences once the Blinn-Phong reflection model is introduced to cater for specular reflection. The third stage is geometry refinement, which refines the initial normal map based on the inferred reflectance maps and refined lighting conditions. The output of the whole process is a 3D face mesh model with a fine (detailed) geometry normal map and two reflectance maps for diffuse and specular albedos. For subsequent new two-view time frames, we use the previously inferred reflectance maps in fitting to newly updated facial geometry together with a refined normal map, and is thus able to adapt to changes in facial expressions.
In the following sections, we will describe the three stages in detail.

\section{Multi-View 3D Face Model Fitting}
Given a pair of unconstrained images from two views, we calculate a 3D face mesh $S$, an initial diffuse albedo map $C_d$, and the rigid face pose $\{R,t,s\}$ denoting the rotation, translation and scale, respectively. In this paper we only consider two views, but the method can be easily extended to multiple views. The fitting module is designed upon the 3DMM model~\cite{blanz19993DMM}, similar to ~\cite{Guo20173DFaceNet,thies2016face}, but with two-view images as input. Once the 3D face mesh is created, we parameterize it to a UV map using a common parametrization method. Then we generate a normal map of the 3D face mesh in the UV map domain and project the diffuse albedo of each visible view into the UV domain as well. We explain the fitting module below.

\subsection{Illumination and shading}
Many previous reconstruction works~\cite{thies2016face,Guo20173DFaceNet,DBLP:journals/corr/SaitoWHNL16} assume surfaces to be Lambertian. To simplify the computation, in the initial stage we also follow assumption and temporarily ignore the specular component.
Thus the shading model for generating facial images in this stage consists of two components:
\begin{align}
I_{ren} = I_{a} + I_{d}
\label{equ:ren}
\end{align}
where $I_a$ and $I_d$  represent the ambient and diffuse components.

We also approximate the global illumination using second order spherical harmonic (SH) functions~\cite{Ramamoorthi:2001:ERI:383259.383317}. The
diffuse component can be expressed as:
\begin{equation}
I_d(p,n,C_d) = C_d(p) \cdot \sum_{k=1}^{B^2} l_k \phi_k (n(p))
\label{equ:eq_diffuse}
\end{equation}
where $p$ denotes the pixel, $C_d(p)$ is the diffuse albedo, $n(p)$ is the normal vector on the 3D facial surface corresponding to the pixel $p$ in the UV space, $\phi$ are the SH basis functions, and $l_k$ are the SH coefficients to be estimated. We use the first $B=3$ bands for the global illumination.

\subsection{Parametric 3D face model} 
We represent 3D face geometry and albedo using a 3D Morphable Model (3DMM) as the parametric face model with z=30k vertices and 59k faces.The 3D face shape geometry $S$ and diffuse colour albedo $C_d$ is compactly encoded via PCA: 
\begin{align}
S &= \bar{S}+A^{id}x ^{id} + A^{exp}x ^{exp} \\
C_d &= \bar{C_d}+A^{alb}x^{alb}
\label{equ:eq_3dmm}
\end{align}
where $\bar{S}$ and $\bar{C_d}$ denote respectively the 3D shape and diffuse albedo of the average face, $A^{id}\in R^{ 3z \times 100} $ and $A^{alb} \in R^{ 3z \times 100}$ are the principal axes extracted from a set of textured 3D meshes with a neutral expression from the Basel Face Model (BFM)~\cite{paysan20093d}, $A^{exp} \in R^{3z \times 79}$ represents the principal axes trained on the offsets between the expression meshes and the neutral meshes of individual persons from FaceWarehouse~\cite{cao2014facewarehouse}, and $x^{id} \in R^{100} $, $x^{exp} \in R^{79}$ and $x^{alb} \in R^{100}$ are the corresponding coefficient vectors that characterize a specific 3D face model.

\subsection{3D face model fitting}
In fitting a 3D face model to 2D views, the unknown parameters to be estimated are $X = \{x^{id},~x^{alb},~x^{exp}, R,t,s,l,I_a\}$. $I_a$ is a scalar value for ambient colour.  We achieve this by minimizing the following objective function:
$$E(X) = E_{con} (X)+ w_l E_{lan}(X)+w_r E_{reg}(X)$$
where $w_l, w_r$ are trade-off parameters balancing the photo-consistency term $E_{con}$, the landmark term $E_{lan}$, and the regularization term $E_{reg}$.We set $w_l = 10$ and $w_r = 5\cdot 10^{-5}$ in our experiments.

The photo-consistency term $E_{con}$ expresses the desire to minimize the difference between synthetic face images and the input images. Since we have two view images as input, we formulate the term $E_{con}$ in ~\cite{thies2016face} to include these two views:
$$E_{con}(X) =\sum_{view}\biggl( \frac{1}{|M_{view}|} \sum_{p \in M_{view}} \Vert I_{ren} - I_{v} \Vert^2\biggr),$$
where $I_v$ is one view of the real face and $I_{ren}$ is the corresponding rendered face from Eq.\ref{equ:ren}, $p \in M_{view} $ denotes a facial pixel in facial pixel set $M_{view}$ from the view image, and $view$ represents the left or right view.

The landmark term $E_{lan}$ encourages the projected facial key vertices of the shape to be close to the corresponding detected landmarks. It is formulated as
$$ E_{lan}(X) =\sum_{view}\biggl( \frac{1}{|F_{view}|} \sum_{f_i \in F_{view}} \Vert f_{i} - \Pi_{view}(R{S_i}+t) \Vert^2\biggr),$$
where $f_{i} \in F_{view}$ is a 2D facial detected landmark in each view using the method of \cite{bulat2017far}, and $\Pi_{view}$ denotes the projection from the 3D face onto the corresponding view image plane. Using two view images improves the robustness of the fitting process, especially in the situation with large pose variation.

The regularization term $E_{reg}$ penalizes facial parameters that are outliers, based on the normal distribution implied by PCA:
$$E_{reg}(X) = \sum_{i = 1}^{100}\bigg[\biggl(\frac{x^{id}_{i}}{\sigma^{id}_i } \biggl)^2 +\biggl(\frac{x^{alb}_{i}}{\sigma^{alb}_i } \biggl)^2  \bigg] +\sum_{i = 1}^{79}\bigg[\biggl(\frac{x^{exp}_{i}}{\sigma^{exp}_i } \biggl)^2\bigg]. $$

We solve the minimization problem by a Gauss-Newton iteration solver implemented in a parallel toolkit~\cite{Cuda}. After solving the $X$ for each pair of input images, we extract the visible view albedo and normal map and project them into the UV texture maps.

\section{Facial Reflectance Inference}
In this section, we infer the specular albedo as well as refine the previously estimated diffuse albedo and lighting, while fixing the facial geometry. To this end, the Blinn-Phong reflectance model~\cite{Blinn:1977:MLR:965141.563893} is used to describe the shading, and an optimization framework is formulated on two-view images.

\subsection{Blinn-Phong reflection model}
\begin{figure}[ht!]
	\centering
	\includegraphics[width=0.7\linewidth]{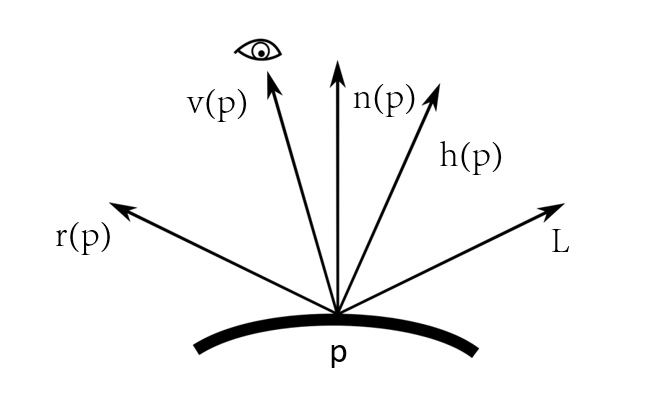}
	\caption{Illustration of the vectors used in the  Blinn-Phong model.}
	\label{fig:Phong_Blinn}
\end{figure}

The Blinn-Phong reflection model is one of common reflectance models that accounts for specular reflection. It computes the specular reflection at $p$ by
\begin{equation}
I_s(p,n,v,C_s) =
\left\{\begin{matrix}
L(p) C_s(p) (n(p) \cdot  h(p) )^m,~n\cdot h> 0\\
0,~~~~~~~~~~~~~~n \cdot h\leq 0
\end{matrix}\right.
\label{equ:eq_specular}
\end{equation}
where $L(p)$ is the incident lighting intensity, $C_s(p)$ is the specular albedo,  $h(p) =\frac{L+v(p)}{|L+v(p)|}$ is a halfway vector between the normalized view vector $v(p)$ towards the viewer(camera) and the normalized lighting vector $L$ towards the light source (see Figure~\ref{fig:Phong_Blinn}), and $m$ is a shininess constant for material, which controls the concentration of the specular reflection.

For the lighting represented by spherical harmonic functions, we use Sloan's method \cite{Shtricks} to extract the dominant lighting direction. The local lighting intensity can be calculated by dividing Spherical Harmonics function value with the product of the dominant direction and the surface normal.

Eq.~\ref{equ:ren} can then be combined with the specular component to give the total reflection of:
\begin{align}
I_{total} = I_{a} + I_{d} + I_{s}
\label{equ:ren_sp}
\end{align}
where $I_a$, $I_d$ and $I_s$ are the ambient, diffuse and specular reflection intensities, respectively.

\subsection{Optimization}
To extract the specular albedo, we introduce a scalar map $C_s$ which stores specular albedos. To refine the diffuse albedo, we introduce a displacement diffuse map $\delta _{C_d}$ that is an offset for the initial diffuse albedo map ${C_d}$.
Then we estimate the displacement diffuse map and the specular albedo map together with the lighting for a subject cross time $t$. This is done by finding $Y =\{\delta_{C_d}, C_s,I_a,l\}$ with $ l= \{l_k\} $ representing the SH coefficients of the lighting,
which is the solution to the following problem:
\begin{equation}
\begin{array}{c}
\underset{Y}{\arg\min}~E_O+\lambda_S E_S+\lambda_H E_H \\ %
 \text{subject~to~} ~0 \leq C_d+\delta_{C_d} \leq 255, ~ 0 \leq C_s \leq 255
\end{array}
\label{equ: objective}
\end{equation}
where $E_O$, $E_S$ and  $E_H$ are the data term, specular difference term and smoothness term, respectively,
while $\lambda_S$ and $\lambda_H$ are two trade-off factors balancing the terms. Below we explain $E_O$, $E_S$ and  $E_H$ in detail.

\paragraph{\textbf{Data term}} The rendering equation of Eq.~\ref{equ:ren_sp} can be re-written with specular albedo $C_s$ and new diffuse albedo $C_d +\delta_{C_d}$:
$$I_{total}(Y,p) = I_a + I_d(C_d +\delta_{C_d},l,p ) +I_s(C_s,l,p)$$
The data term $E_O$ measures the difference between the rendered image $I_{total}(Y,p)$ and the corresponding observed images $I_{t,view}$ of each view  at time $t$:
\begin{align}
	\begin{split}
	E_{O}(Y) = \sum_{t} \sum_{view}\sum_{p \in M_{view}} \Vert I_{t,view}(p) -I_{total}(Y,p) \Vert _{2}^2
	\end{split}
\end{align}

All images of the same subject are recorded in a consistent environment. Thus the ambient and lighting conditions are estimated globally for all images.

\paragraph{\textbf{Specular difference loss term}}
Since ambient and diffuse reflections are view independent, the image difference between the two views is thus mainly caused by the difference in specular reflection. We use this property to estimate the specular reflection in an overlapping region of the two views. To get a full face specular albedo map, we construct the objective function from several pairs of $(I_{l,t} ,I_{r,t}) $ obtained across different timestamps $t$. We aim to minimize the specular difference loss  $\|\widetilde{ I_t} - \widetilde{ I_{s,t}}\|$, where  $\widetilde{ I_t} = I_{l,t} - I_{r,t}$ is the difference between the captured images in the two views, while $\widetilde{ I_{s,t}} = I_s(p,n_t,v_{l,t},C_s) -I_s(p,n_t,v_{r,t},C_s)$ is the difference between the rendered images in the two views, with the left and right view vectors given by $v_{l,t}$ and $v_{r,t}$ respectively. The view vector maps can be obtained after the 3D face model fitting described in the previous section. The overall loss is thus written as follows:
$$ E_S(C_s,l) = \sum_{t} \sum_{p \in M_{r}\cap  M_{l}} \Vert \widetilde{I_t}(p) - \widetilde{I_{s,t}}(p,l,C_s(p)) \Vert^2 $$

\paragraph{\textbf{Smoothness term}} To constrain the gradients of the diffuse displacements as well as the specular albedoes to be locally smooth, the smoothness term $E_H$ is introduced:
$$ E_H(\delta_{C_d},C_s) =\sum_{p}(\Vert \Delta \delta_{C_d(p)} \Vert_1 + \Vert \Delta C_s(p) \Vert_1), $$
where $\Delta$ is the Laplacian operator. Here the $l_1$ norm is used to better preserve sharp discontinuities.

To solve the minimization problem of Eq.~\ref{equ: objective}, the parameters are estimated sequentially. We first update lighting $l$ and ambient colour $I_a$, followed by the specular albedo map $C_s$, and finally the diffuse displacement $ \delta_{C_d}$.
The optimization is iterated until convergence. We set the trade-off parameters $\{\lambda_S, \lambda_H \}$ to be $\{0.1,~0.001\}$ and empirically set the shininess exponent $m$ to be 5 in our experiment. For timestamps, we choose 3 or 4 pairs of frames of a subject with different poses. The initial diffuse albedo, SH coefficients and ambient scalar is set as mean value derived from $X$ of all frames solved in the previous section. The initial specular is set as zero map.  The energy function is optimized by the Adam\cite{DBLP:journals/corr/KingmaB14} solver in parallel.

\section{Geometry Refinement}
Once facial reflectance inference is completed, we proceed to refine the normal map obtained from the multi-view facial model fitting. For this purpose, we introduce a correction map $\delta_n$ that is used to correct the normal vector stored in the normal map to improve the matching of the rendering images and input images. The correction vector is computed by minimizing the following objective function:
\begin{align}
E(\delta_n ) =  \sum_{view} \sum_{p \in M_{view}} \Vert I_{view}(p) -  I_{total}(n(p)+\delta_n(p)) \Vert^2 \nonumber \\
+ w_1 \Vert \Delta \delta_n \Vert _{1} +w_2  \Vert \delta_n \Vert _{2}^2,
\label{equ:loss}
\end{align}
where $I_{total}(n(p)+\delta_n(p))$ is the image generated by the rendering equation Eq.~\ref{equ:ren_sp} with the new normal vector $n(p)+\delta_n(p)$, the term $\Vert \delta_n \Vert _{2}^2$ is to penalize large perturbation, the Laplacian term in $l_1$ norm, $\Vert \Delta \delta_n \Vert _{1}$, is a smoothness term but allowing the existence of sharp discontinuities, and $w_1$ and $w_2$ are the trade-off parameters. We set $w_1$ and $w_2$ to be ${10^{-3}}$ and 0.3 in our experiment, and use spherical coordinates to represent normals for a smoother space. The minimization problem is solved using the Adam\cite{DBLP:journals/corr/KingmaB14} solver in parallel.

Once normal map is refined, we can update vertex positions of the facial mesh to improve the geometry by a least-squares process.

Note that the diffuse and specular albedos are independent of time and view. Once they are computed, they remain the same for new coming images. Thus for a new coming pair of images whose geometry may change due to different facial expressions or other factors, we only need to recover the facial geometry. This can be done by our proposed procedure but with fixed diffuse and specular albedos, which greatly simplifies the computation.

%% file: experiment.tex
\section{Experiments}

We recorded videos from three views for each subject: two views are used as input to recover the facial reflectance and geometry, and the third view is used as the ground truth view for evaluation. All the videos were captured simultaneously at 30 FPS with 680x480 resolution. For synchronization, each camera is controlled by an individual PCI port with UDP communication module.

More specifically, we recorded videos of 15 persons (12 male and 3 female) with different ethnicities. Each subject was required to rotate his or her head and perform different facial actions based on the Facial Action Coding System(FACS)~\cite{ekman_1978}, such as opening and closing the mouth, and various expressions. We used the first few frames to infer the reflectance model, which includes the global lighting, ambient colour, diffuse albedo texture and specular albedo. For the remaining frames, we obtained the initial geometry normals and further refined the normal directions to recover subtle skin deformation caused by facial movement. Finally, we used the reflectance model and the refined geometry normals to render the face in the ground truth viewpoint, for purposes of evaluation and comparison.

\begin{figure}
	\centering
	\includegraphics[width=1.0\linewidth]{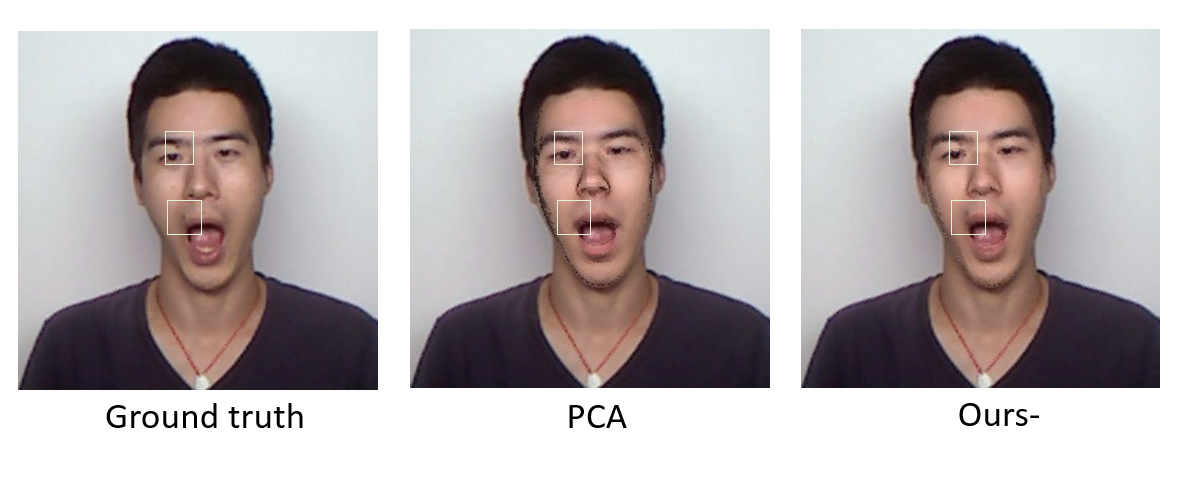}
	\caption{Comparison of our method with the PCA method in recovery of diffuse albedo. ``Ours-'' is rendered with our diffuse albedo. ``PCA'' is rendered using diffuse albedo obtained by the PCA method, where the appearance feature like mole is missing in the rectangle regions.}
	\label{fig:pca}
\end{figure}

\begin{figure*}
	\centering
	\includegraphics[width=0.8\linewidth]{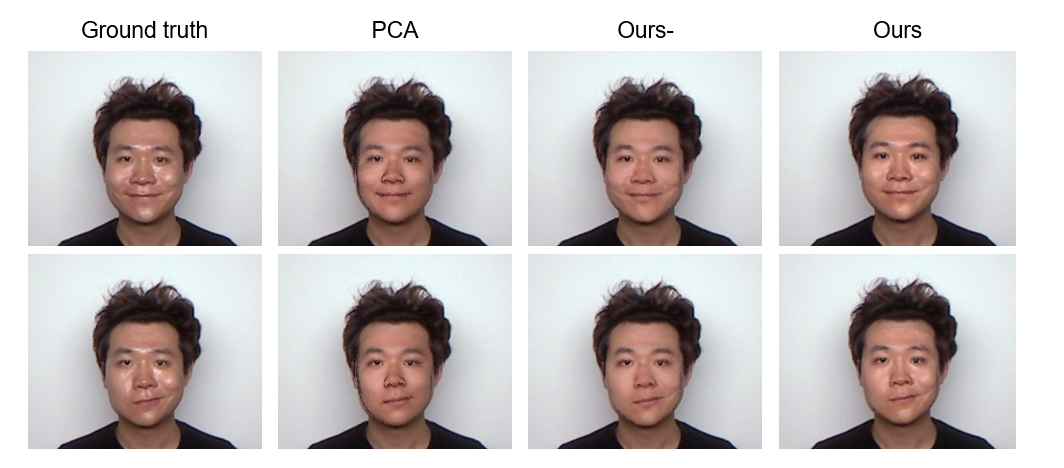}
	\caption{Comparison in recovering geometry. From left to right are the ground truth image, diffuse colour rendering using geometry normal recovered by the PCA based method~\cite{Blanz:1999:MMS:311535.311556}, diffuse colour rendering (Ours-) using our refined geometry normal, and diffuse and specular colour rendering (Ours) using our refined geometry normal. }
	\label{fig:specu}
\end{figure*}

\paragraph{\textbf{PCA vs Ours}}
The PCA-based Lambertian model (Section 4.3) is widely used in 3D facial reconstruction. Figure~\ref{fig:pca} shows the diffuse albedo recovered by our method and by the PCA-based Lambertian model. It can be seen that our diffuse albedos reveal high frequency skin features such as moles, while the PCA model only recovers a low frequency diffuse albedo map that is unable to capture the fine details. We further compared the geometry refinement in Figure~\ref{fig:specu}. The PCA-based approach uses a low dimension basis to represent geometry, while ours is able to further refine the mesh normals so as to better capture detailed surface variation (e.g., regions around the mouth).

\begin{figure}[h!]
	\centering
	\includegraphics[width=1.0\linewidth]{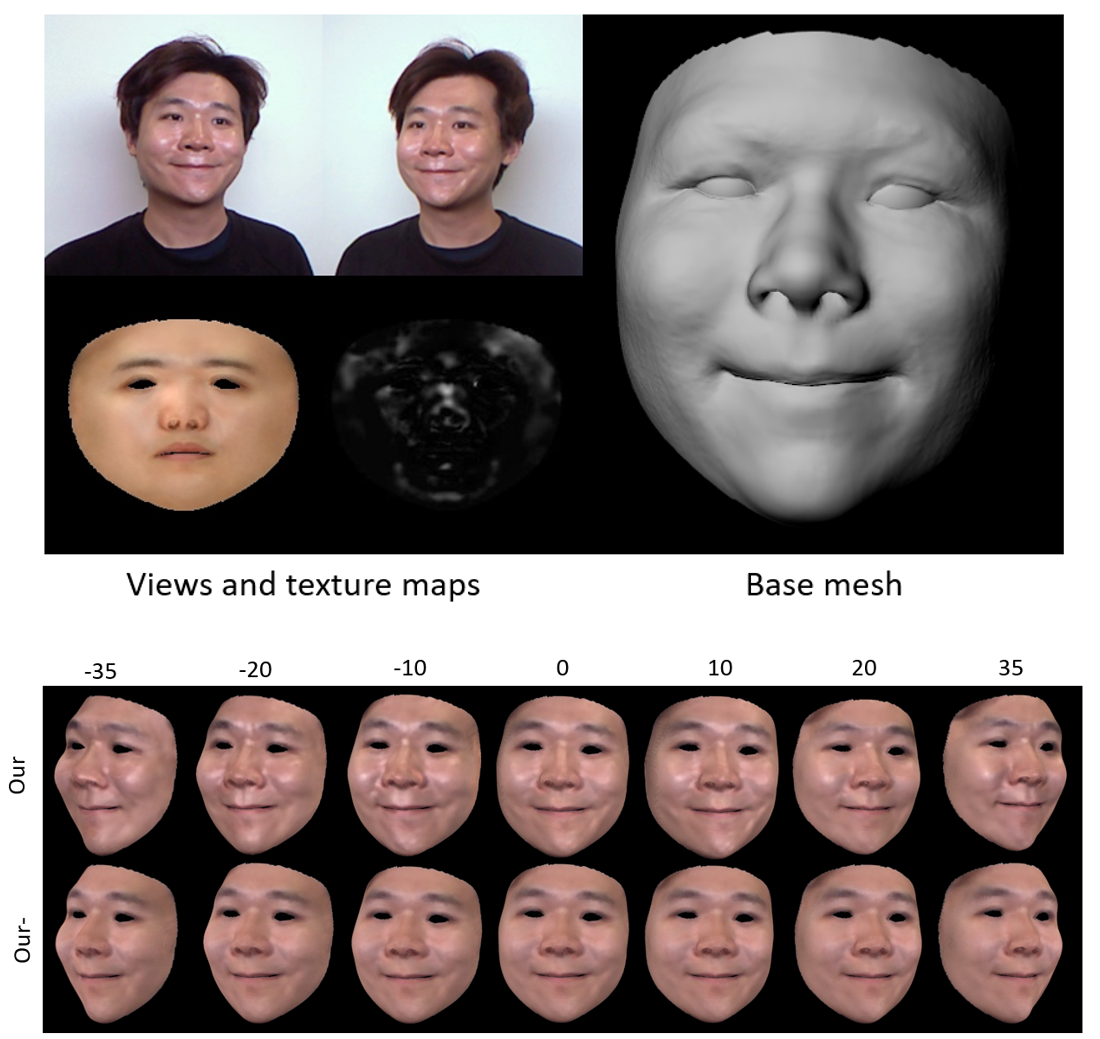}
	\caption{Rendering results of different view angles. The top images show the two view inputs with inferred reflectance textures and the corresponding fitted base mesh result. The bottom images show the rendering faces (with refined normal) when a virtual camera rotates about a vertical axis from -35$^\circ$ to 35$^\circ$: the first row is rendered with specular effect and the second row is rendered without specular effect.}
	\label{fig:synthetic_face}
\end{figure}

\paragraph{\textbf{Specular effect}}
Figure~\ref{fig:synthetic_face} shows that when a pair frames and inferred reflectance textures are given, our method can recover photorealistic faces with specular effects in new viewpoints. Compared to the view-independent Lambertian model, our approach can synthesize more plausible results with view changes or facial movement. More results can be found in the supplementary video.

\subsection{Face tracking comparison}
\begin{figure}
	\centering
	\includegraphics[width=1.0\linewidth]{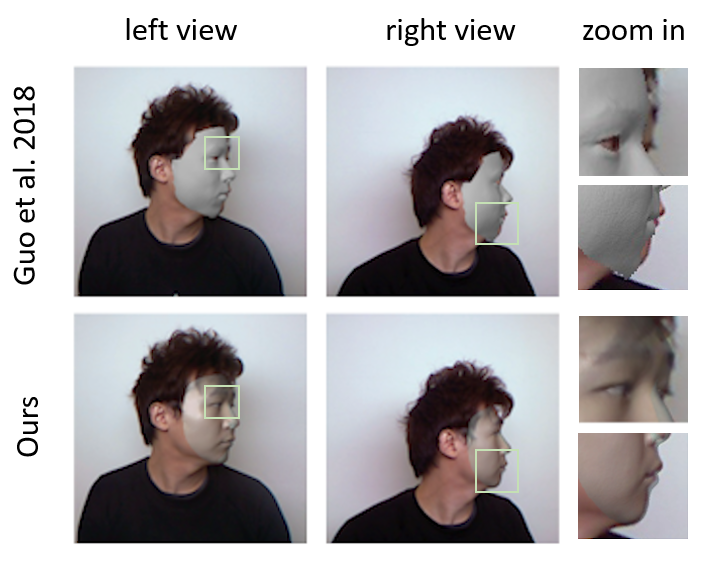}
	\caption{Comparison of 3D face fitting with CoarseNet of Guo et al.\cite{Guo20173DFaceNet} in an extreme pose. From left to right are the fitting overlap results on the left view, right view and zoom in detail of green window. Note that Guo et al~\cite{Guo20173DFaceNet}  performs fitting twice for two views and our method performs the fitting simultaneously for both views.}
	\label{fig:tracking_compare}
\end{figure}

While face tracking is not the focus of our work, Figure~\ref{fig:tracking_compare} shows that our multi-view 3D face tracking can handle extreme facial poses compared to the state of the art single view method. Guo et al.~\cite{Guo20173DFaceNet} use network to regress 3DMM parameters and reconstruct 3D meshes from an RGB images. In most cases, both methods produce similar high quality results (that can be found in the supplementary video). The closeups in Figure~\ref{fig:tracking_compare} show that our method yields a closer face fit in an extreme pose, which is particularly visible at the silhouette of the input face. In general our multi-view fitting can handle more challenging poses and is more stable than the methods using only monocular input.

\subsection{Capture performance comparison}

\begin{figure*}
	\centering
	\includegraphics[width=0.95\linewidth]{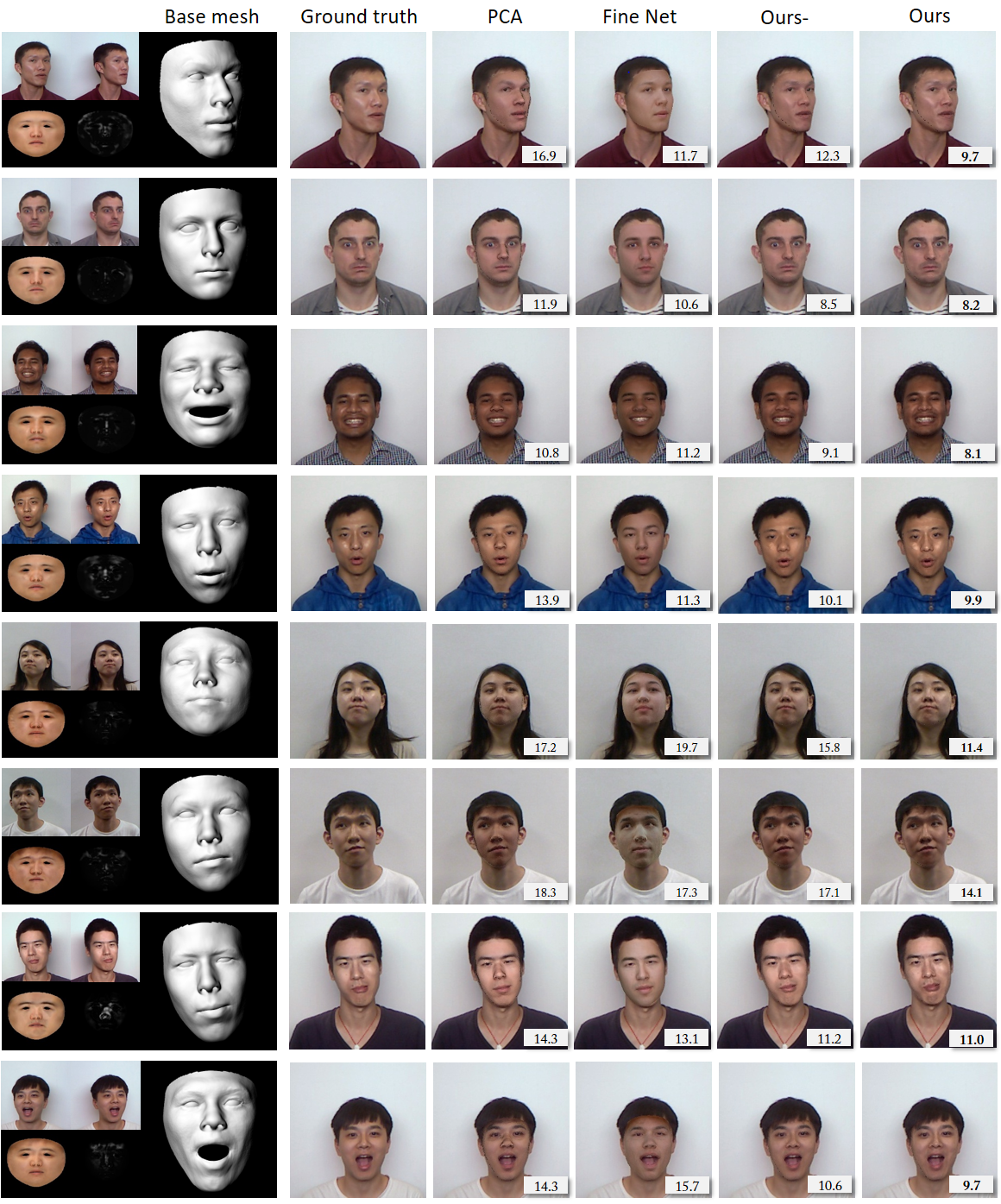}
	\caption{Rendering results of different subjects. On the left side are the input two view frames, recovered reflectance maps and fitted base mesh projected to the ground truth view. On the right side are ground truth view and overlapped facial rendering results obtained from PCA based method~\cite{Blanz:1999:MMS:311535.311556}, FineNet~\cite{Guo20173DFaceNet}, our method without specular effect, and our method with specular effect. The RMSE is given in the lower-right corner of each rendered image.}
	\label{fig:gt}
\end{figure*}

In Figure~\ref{fig:gt}, we qualitatively compared our approach with classic PCA-based 3D face reconstruction~\cite{Blanz:1999:MMS:311535.311556} and the state-of-the-art CNN-based FineNet~\cite{Guo20173DFaceNet} for 8 subjects. The PCA-based method reconstructs 3D faces and textures using regressed coefficients for a low dimensional basis, and has difficulty in synthesizing skin appearance details like wrinkles and mode. FineNet~\cite{Guo20173DFaceNet} first regresses coarse 3D faces from input images and then refines the 3D face geometry details, but it does not refine texture. Moreover, both PCA and FineNet assume Lambertian surfaces and cannot render specular effects. Our method handles geometry, texture refinement and specular effects, which leads to more photorealistic results. A video comparison can be found in the supplementary video.

In Table~\ref{tab:quantitative}, we quantitatively measured the ability of our approach to faithfully render photorealistic faces. We tested 1500 frames from two views of 15 subjects with different expressions and synthesized to a third view, for which we had the corresponding ground truth images. Compared to the PCA based method~\cite{Blanz:1999:MMS:311535.311556} and FineNet~\cite{Guo20173DFaceNet}, our approach obtained the highest score in PSNR and the lowest error in RMSE.
\begin{table}
	\caption[Caption for LOF]{Quantitative results of face rendering, measured using the PSNR and the root-mean-square error (RMSE) in the same overlap facial region\protect\footnotemark\ . }
	\begin{tabular}{c|cc}
		Methods &PSNR & RMSE\\
		\hline\hline
		PCA~\cite{Blanz:1999:MMS:311535.311556} &24.6 &15.0\\
		FineNet~\cite{Guo20173DFaceNet} &25.5 &13.6  \\
		Our method (without specular)&26.8&11.6  \\
		Our method (with specular) &28.3 &9.8  \\
	\end{tabular}
	\label{tab:quantitative}
\end{table}
\footnotetext{The facial mesh in FineNet is larger than ours. For fair comparison, we calculate the error only within the same facial region from our face projection.}

\subsection{Timing}
All our experiments were conducted on a PC with Intel Core i7-7700K CPU of 3.5 GHz equipped with two GeForce GTX 1080 with 8GB memory. Following the pipeline in Figure~\ref{fig:overview}, our multi-view 3D face model fitting takes less than 2 seconds, the reflectance inference consumes about 42 seconds for one subject, and the geometry refinement with 100 iteration takes 2 seconds per pair views.